\documentclass[sigplan,screen]{acmart} % ,review,anonymous
\AtBeginDocument{%
  }

\setcopyright{acmlicensed}
\copyrightyear{2025}
\acmYear{2025}
\acmDOI{XXXXXXX.XXXXXXX}
\acmISBN{978-1-4503-XXXX-X/2018/06}

\begin{document}

\title{ZSE-Cap: A Zero-Shot Ensemble for Image Retrieval and Prompt-Guided Captioning}


\author{Duc-Tai Dinh}
\email{ductai.dt05@gmail.com}
\orcid{0009-0002-4958-8960}
\authornotemark[1]
\affiliation{%
  \institution{University of Science, VNU-HCM}
  \city{Ho Chi Minh City}
  \country{Vietnam}
}

\author{Duc Anh Khoa Dinh}
\authornote{Both authors contributed equally to this research.}
\email{dinhducanhkhoa@gmail.com}
\orcid{0009-0001-8569-1976}
\affiliation{%
  \institution{University of Science, VNU-HCM}
  \city{Ho Chi Minh City}
  \country{Vietnam}
}

\renewcommand{\shortauthors}{Dinh et al.}

\begin{abstract}
We present ZSE-Cap (Zero-Shot Ensemble for Captioning), our 4th place system in Event-Enriched Image Analysis (EVENTA) shared task on article-grounded image retrieval and captioning. Our zero-shot approach requires no fine-tuning on the competition's data. For retrieval, we ensemble similarity scores from CLIP, SigLIP, and DINOv2. For captioning, we leverage a carefully engineered prompt to guide the Gemma 3 model, enabling it to link high-level events from the article to the visual content in the image. Our system achieved a final score of 0.42002, securing a top-4 position on the private test set, demonstrating the effectiveness of combining foundation models through ensembling and prompting. Our code is available at \url{https://github.com/ductai05/ZSE-Cap}.
\end{abstract}



\begin{teaserfigure}
  \includegraphics[width=\textwidth]{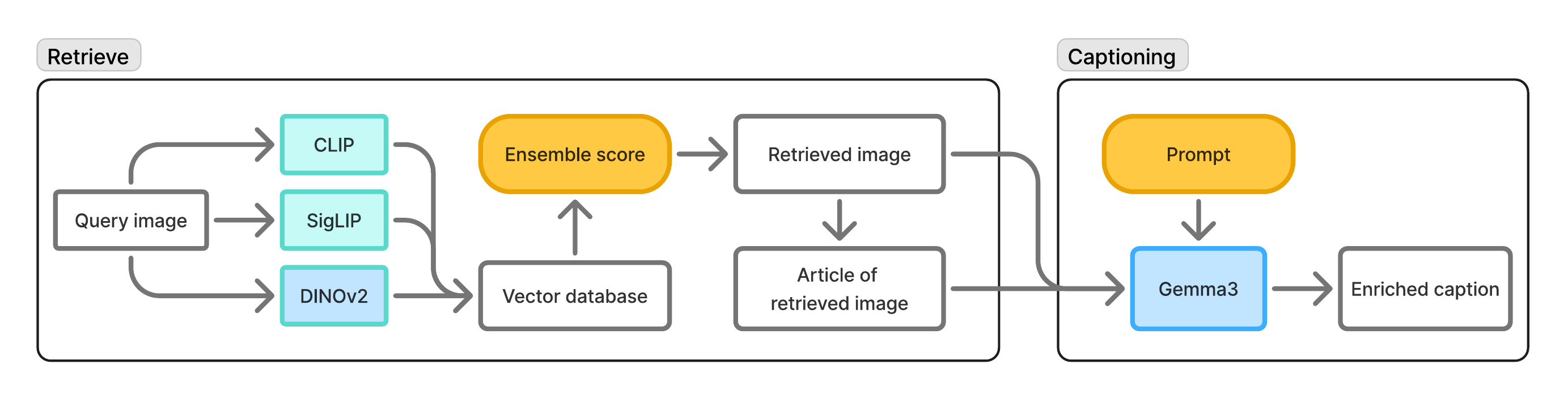}
  \caption{ZSE-Cap's pipeline for image retrieval and captioning}
  \Description{ZSE-Cap's pipeline: Query image -> (CLIP, SigLIP, DINOv2) -> Gemma3 -> Caption}
  \label{fig:teaser}
\end{teaserfigure}


\maketitle

% ZSE-Cap
\section{Introduction}
The exponential growth of digital content has created a pressing need for intelligent systems that can bridge the gap between long-form text and visual media. While significant progress has been made in generic image search and description, real-world applications often demand a deeper, context-aware understanding. A prime example is connecting a news article to its corresponding images, which requires not only identifying relevant visuals but also describing them in a way that reflects the article's specific narrative and events.

The Event-Enriched Image Analysis (EVENTA - Track 1: Event-Enriched Image Retrieval and Captioning) shared task, introduced with the OpenEvents V1 dataset (\citet{nguyen2025openevents}), formalizes this challenge by proposing two coupled sub-tasks: article-grounded image retrieval and captioning. As described by Nguyen et al., this task pushes the boundaries of current multi-modal systems by requiring them to parse the complex semantics of a full article to guide both search and generation. The core difficulty, highlighted by the dataset's authors, lies in accurately linking high-level textual concepts (e.g., political events, specific individuals' actions) to low-level visual features within a vast image corpus of over 400,000 images sourced from news outlets.

Faced with this challenge, we diverge from a conventional fine-tuning approach—which risks overfitting and demands significant computational resources—to propose a novel zero-shot pipeline, \textbf{ZSE-Cap}. Our philosophy is that intelligently combining the collective capabilities of existing foundation models is a more robust and scalable strategy than task-specific training. For retrieval, we employ an ensemble method that combines signals from the CLIP, SigLIP, and DINOv2 models to ensure robust and accurate image selection. For captioning, we utilize the powerful Gemma 3 model, guided by a carefully engineered prompt that directs it to perform a complex reasoning task: linking the article's key events to the visual content of the retrieved image.

Our system, which secured a top-4 finish in the competition, makes the following contributions:

\begin{itemize}
    \item A novel and effective zero-shot pipeline for the joint task of article-grounded image retrieval and captioning, validated by a top-4 rank in a competitive shared task.

    \item A robust retrieval module based on ensembling scores from diverse vision models (CLIP, SigLIP, DINOv2), which proves more effective than relying on a single model.

    \item A demonstration of prompt-guided captioning, where a state-of-the-art LLM (Gemma 3) is successfully steered to generate contextually rich captions that link article events to visual evidence.

    \item ZSE-Cap's code: \url{https://github.com/ductai05/ZSE-Cap}
\end{itemize}
\section{Related Work}

Our approach is informed by advancements across several key domains. This section first reviews the landscape of context-grounded multimodal benchmarks, establishing the problem setting for our work. We then discuss the core technologies our system integrates: vision-language models for retrieval, self-supervised visual learning, and large-scale language models for prompt-guided generation.

\subsection{Context-Grounded Multimodal Benchmarks}

While foundational datasets like MS-COCO \cite{lin2014microsoft} or Flickr30k \cite{plummer2015flickr30k} have driven progress in general image captioning, they often focus on object-level descriptions. Recently, there has been a push towards creating benchmarks that require deeper, contextual understanding. The \textbf{OpenEvents V1} (\citet{nguyen2025openevents}) dataset is a prime example, specifically designed for event-centric grounding. It provides over 200,000 news articles and 400,000 images, challenging models to move beyond surface-level descriptions and connect visual content to real-world narratives. Unlike traditional retrieval tasks, the queries in OpenEvents V1 are event-rich descriptions, making simple visual-semantic matching insufficient. The EVENTA shared task, in which we participate, is built upon this very benchmark, providing a standardized protocol to evaluate systems on this complex, context-aware task. Our work directly addresses the challenges proposed by this benchmark.

\subsection{Vision-Language Pre-training for Retrieval}
The dominant paradigm for image-text retrieval is built on joint vision-language embedding spaces. 
\citet{radford2021learning} pioneered this approach with \textbf{CLIP} (Contrastive Language-Image Pre-training), training a dual-encoder architecture on 400 million image-text pairs using a contrastive loss. This enables a powerful zero-shot transfer capability. 
More recently, \textbf{SigLIP} (Sigmoid-based Language-Image Pre-training) has emerged as a strong alternative \textbf{\cite{zhai2023sigmoid}}, which has been shown to improve performance and training stability. 
Our work leverages both CLIP and SigLIP, hypothesizing that their different training objectives provide complementary signals for our retrieval ensemble.

\subsection{Self-Supervised Learning for Vision}
Beyond models trained with explicit text supervision, self-supervised learning has produced powerful vision-only encoders. 
\textbf{DINOv2 \cite{oquab2023dinov2}} is a state-of-the-art example, learning rich visual representations by applying self-distillation to a massive, curated dataset of images without any textual labels.

\subsection{Large Language Models for Generation and Reasoning}
The task of image captioning has evolved from earlier encoder-decoder architectures to leveraging the immense power of Large Language Models (LLMs). 
Models like \textbf{Gemma 3 \cite{team2025gemma}}, a family of lightweight, state-of-the-art open models from Google, demonstrate strong text generation and instruction-following capabilities. Our work adopts this strategy to steer Gemma 3 beyond simple description.
\section{Methodology}

Our proposed system, ZSE-Cap, is designed as a two-stage, zero-shot pipeline that requires no fine-tuning on the competition's training data. The first stage performs a robust image-to-image retrieval to find the relevant context article, and the second stage generates a descriptive caption for the original query image based on this retrieved context. The complete pipeline is illustrated in figure \ref{fig:teaser}.

\subsection{Stage 1: Ensemble-based Image-to-Image Retrieval}
The primary objective of this initial stage is to identify the most relevant news article from the corpus that corresponds to a given query image, $I_q$. Our approach is predicated on the hypothesis that ensembling multiple diverse foundation models can yield a more robust and accurate retrieval performance than any single model, as individual models may capture different facets of visual information.

To this end, we selected three distinct and powerful pre-trained models: CLIP, SigLIP, and DINOv2. We selected CLIP, SigLIP, and DINOv2 for their complementary features. CLIP and SigLIP provide rich visual-semantic alignments, while the self-supervised DINOv2 offers fine-grained visual pattern recognition. This combination allows matching based on both high-level concepts and detailed visual patterns.

For the ensemble mechanism, we first pre-compute the feature embeddings for every image in the database using all three models. Given a query image, we extract its corresponding embeddings and calculate the L2 distance to every image in the database, generating three separate ranked lists of candidate images. We then investigated two methods to fuse these rankings:

\begin{itemize}
    \item \textbf{Weighted Ensemble:} This method computes a final similarity score based on a weighted sum of L2 distances from each model. For a candidate image $I_c$ in the database, the L2 distance $d_{\text{model}}(I_q, I_c)$ is calculated in each feature space.  The final ensemble score for each candidate image is calculated as:
    $$
        S_{\text{WE}}(I_c) = \sum_{m \in \{\text{CLIP, SigLIP, DINOv2}\}} w_m \cdot d_{\text{m}}(I_q, I_c)
        \label{eq:ensemble}
    $$
    where $d_{\text{m}}$ is the L2 distance for model $m$, and $w_m$ is the assigned weight. Based on empirical evaluation, we assigned weights of 0.5 to DINOv2, and 0.3 to both SigLIP and CLIP. To normalize, each weight was scaled proportionally relative to the total sum of all weights. This resulted in final weights of 0.4545 for DINOv2 and 0.2727 for both SigLIP and CLIP. The candidate image with the lowest $S_{\text{WE}}$ is selected.

    \item \textbf{Reciprocal Rank Fusion (RRF):} This is a parameter-free method that combines rankings from multiple systems. The RRF score for a candidate image $d$ is calculated as:
    $$ \text{S}_{\text{RRF}}(d) = \sum_{r \in R} \frac{1}{k + \text{rank}_r(d)} $$
    where $R$ is the set of rankings from our three models, $\text{rank}_r(d)$ is the rank of image $d$ in ranking $r$, and we set the constant $k=0$. The candidate image with the highest $S_{\text{RRF}}$ is selected.
\end{itemize}
Upon evaluating both strategies on the competition's public test set, the \textbf{weighted ensemble} method demonstrated superior performance. Consequently, it was selected as the definitive retrieval mechanism for our final system, providing the retrieved article context for the subsequent captioning stage.

\subsection{Stage 2: Prompt-Guided Caption Generation}

In this stage, our goal is to generate a caption for the original query image $I_q$ that is contextually grounded in the content of the retrieved article. We leverage the advanced instruction-following and multi-modal reasoning capabilities of the Gemma 3 model in a zero-shot setting.

The core of our approach lies in prompt engineering. We designed a specific prompt structure to explicitly instruct the model to perform a synthesis task: analyze the article's content and use that information to describe the visual elements of the query image. The input provided to the Gemma 3 model is a carefully constructed multi-modal triplet: (1) the Query Image, (2) the full content of the Retrieved Article, and (3) our Engineered Prompt. This structure is deliberate:

\begin{itemize}
    \item The Query Image serves as the primary visual anchor, ensuring the generated caption is factually grounded in what is actually depicted.

    \item The Retrieved Article acts as the external knowledge base, providing the rich, event-centric context (such as names, locations, causes, and consequences) that is often absent from the image itself.

    \item The Engineered Prompt functions as the crucial cognitive orchestrator. It explicitly instructs the model how to synthesize information from the other two inputs. Instead of merely describing the image or summarizing the article, the prompt forces the model to perform a higher-level reasoning task: to find and articulate the direct link between the visual evidence and the article's narrative.
\end{itemize}

This three-part input transforms the task from simple captioning into a sophisticated process of retrieval-augmented synthesis, which is key to our method's success in generating contextually rich and relevant descriptions.

The engineered prompt provided to Gemma 3 is structured as follows:

\begin{quote}

\small 

\vspace{0.5em} 

\ttfamily 

You are a seasoned expert in journalistic photo caption writing. Your primary goal is to create a SINGLE, comprehensive caption that \textbf{masterfully connects the provided IMAGE to the key events, figures, and narratives within the ARTICLE.}

This caption must achieve the following, with a strong emphasis on the connection to the article:

\begin{enumerate}
    \item \textbf{Contextualize the Image through the Article First:}
    \begin{itemize}
        \item Before describing visual details, understand the \textit{core message, main event, or central figures discussed in the ARTICLE.}
        \item Identify \textit{how the IMAGE serves to illustrate, support, or provide a visual anchor for these key aspects of the article.}
    \end{itemize}

    \item \textbf{Describe the Image in Service of the Article's Narrative:}
    \begin{itemize}
        \item Briefly describe the \textit{most relevant visual elements of the IMAGE} -- focusing on subjects, actions, and settings that directly pertain to the article's content.
        \item \textit{Prioritize details that help the reader understand the image's role in the story.} Less critical visual minutiae can be omitted if they don't serve this primary purpose.
        \item If specific named entities (people, places, organizations) from the article are clearly visible and relevant, ensure they are mentioned to strengthen the link.
    \end{itemize}

    \item \textbf{Clearly Articulate the Significance and Connection:}
    \begin{itemize}
        \item The caption's main thrust should be to clarify the \textit{image's significance and its direct relationship to the information presented in the article.}
        \item Ensure the reader understands \textit{why} this specific image accompanies this article.
    \end{itemize}

    \item \textbf{Professional and Informative Style:}
    \begin{itemize}
        \item Use precise, informative, and coherent language.
        \item The style should be journalistic, conveying information efficiently and accurately.
    \end{itemize}
\end{enumerate}

\textbf{Crucial Requirement:} Begin writing the caption IMMEDIATELY. Do not include any introductory phrases such as \texttt{'Here is the caption:'}, \texttt{'The caption is:'}, or similar.
\end{quote}

This structured prompt guides the LLM through a multi-step reasoning process: first, contextualizing the image within the article's narrative (Step 1); second, describing visual elements in service of that narrative (Step 2); and finally, explicitly articulating the connection (Step 3). This prevents the model from generating generic, disconnected descriptions.
\section{Experiments}

This section details the experimental setup of our ZSE-Cap system on the EVENTA shared task.

\begin{itemize}
    \item Dataset: All experiments were conducted using the OpenEvents V1 dataset provided by the competition organizers. We used the \texttt{public\_test} set for development, hyperparameter tuning (specifically, determining the ensemble weights), and ablation studies. The final, official evaluation was performed on the blind \texttt{private\_test} set.

    \item Evaluation Metrics: We adhere to the official metrics of the shared task.
    \begin{itemize}
        \item For Retrieval: Mean Average Precision (mAP), Recall@1 (R@1), and Recall@10 (R@10).
        \item For Captioning: CLIPScore for semantic relevance and CIDEr for consensus-based factual correctness against ground-truth captions.
        \item Computational Environment: Our experiments were primarily conducted on the Kaggle platform, utilizing standard notebooks equipped with NVIDIA T4 (x2), P100 GPUs or TPU VM v3-8. This environment was sufficient for running the feature extraction with CLIP, SigLIP, DINOv2, and for inference with the Gemma-3-4B-it model. Due to the significant memory and computational requirements of the Gemma-3-27B-it model, which exceed the limits of standard Kaggle notebooks, inference for this model was performed on a more powerful environment. This two-pronged approach allowed us to leverage the best-performing models while working within practical resource constraints.
    \end{itemize}
\end{itemize}

\section{Results}

This section details the performance of our ZSE-Cap system. We first present the results and ablation studies conducted on the public test set, which guided our model development. We then report our final, official scores on the private test set. Finally, we provide a qualitative analysis to illustrate the practical impact of our approach.

\subsection{Public test set}

The public test set was our primary environment for development, validation, and hyperparameter tuning. The results on this set were crucial for finalizing our ensemble weights and prompt engineering strategy.

\begin{table}[H]
\centering
\caption{Retrieval results on the public test set}
\label{tab:retrieval-public}
\begin{tabular}{@{}llll@{}}
\toprule
\textbf{Method}        & \textbf{mAP} & \textbf{R@1} & \textbf{R@10} \\ \midrule
BLIP                   & 0.589        & 0.475        & 0.746         \\
CLIP                   & 0.981        & 0.969        & 0.997         \\
SigLIP                   & \multicolumn{1}{c}{-}        & 0.973        & 0.998         \\
DINOV2                   & \multicolumn{1}{c}{-}        & 0.971        & 0.998         \\
CLIP + SigLIP + DINOv2 (WE) & 0.994        & 0.99         & 0.999         \\ 
CLIP + SigLIP + DINOv2 (RRF) & 0.991        & 0.985         & 0.999         \\ \bottomrule
\end{tabular}
\end{table}

\begin{table}[H]
\centering
\caption{Captioning results on the public test set}
\label{tab:captioning-public}
\begin{tabular}{@{}lll@{}}
\toprule
\textbf{Method}                  & \textbf{CLIPScore} & \textbf{CIDEr} \\ \midrule
gemma-3-4b-it + Image             & 0.820              & 0.001          \\
gemma-3-4b-it + Article + Prompt  & 0.817              & 0.133          \\
gemma-3-27b-it + Article + Prompt & 0.842              & 0.151          \\ \bottomrule
\end{tabular}
\end{table}

\textbf{Retrieval Performance}: To validate our ensemble approach, we benchmarked the full ZSE-Cap system against its individual component models. Table \ref{tab:retrieval-public} presents the retrieval performance.

The results clearly validate our ensemble strategy. While CLIP demonstrates impressive performance on its own, our weighted ensemble, ZSE-Cap, consistently outperforms it across all metrics. Notably, the ensemble achieves an mAP of 0.994 and an R@1 of 0.99. This demonstrates that by fusing the fine-grained visual patterns from DINOv2 with the rich visual-semantic alignments from CLIP and SigLIP, our system achieves a more robust and comprehensive image understanding, leading to superior retrieval accuracy.

\textbf{Captioning Performance}: Table \ref{tab:captioning-public} details the captioning results, comparing different configurations of our generation module.

The impact of our article-grounded prompting is dramatic. A baseline using only the image yields a CIDEr score of 0.001; providing the article and prompt boosts this to 0.133, a 130-fold increase. This demonstrates our method successfully steers the LLM to synthesize event-specific details from the text. This 130-fold increase underscores that our method successfully steers the LLM to synthesize event-specific details (such as names, locations, and context) from the text, which are essential for aligning with the ground-truth captions. Furthermore, scaling the model to gemma-3-27b-it further boosts performance to a CLIPScore of 0.842 and a CIDEr of 0.151, confirming the benefits of a more capable language model for this complex reasoning and synthesis task.

\subsection{Private test set}

The private test set served as the final, unseen benchmark for the EVENTA shared task, with the scores on this set determining the official rankings. Our system was evaluated in its final configuration: the weighted ensemble for retrieval and the Gemma-3-27b-it model for captioning.

\begin{table}[H]
\centering
\caption{Retrieval results on the private test set}
\label{tab:retrieval-private}
\begin{tabular}{@{}llll@{}}
\toprule
\textbf{Method}        & \textbf{mAP} & \textbf{R@1} & \textbf{R@10} \\ \midrule
CLIP + SigLIP + DINOv2 (WE) & 0.966        & 0.955        & 0.983         \\ \bottomrule
\end{tabular}
\end{table}

\begin{table}[H]
\centering
\caption{Captioning results on the private test set}
\label{tab:captioning-private}
\begin{tabular}{@{}lll@{}}
\toprule
\textbf{Method}                  & \textbf{CLIPScore} & \textbf{CIDEr} \\ \midrule
Gemma3-27b-it + Article + Prompt & 0.828              & 0.133          \\ \bottomrule
\end{tabular}
\end{table}

Tables \ref{tab:retrieval-private} and \ref{tab:captioning-private} present the final, official scores of ZSE-Cap. Our system achieved a final score of 0.42002, securing a top-4 rank in the competition.

For the retrieval sub-task, ZSE-Cap achieved a robust mAP of 0.966 and R@1 of 0.955. For the captioning sub-task, it obtained a CLIPScore of 0.828 and a CIDEr score of 0.133. While there is a slight performance moderation compared to the public set—a common occurrence when transitioning to blind test data—the results remain highly competitive. This strong performance on the final test set validates the generalization capability of our zero-shot approach, proving its effectiveness in a real-world scenario without any task-specific fine-tuning.

\subsection{Qualitative analysis}

\begin{figure}[H]
\centering
\includegraphics[width=1\linewidth]{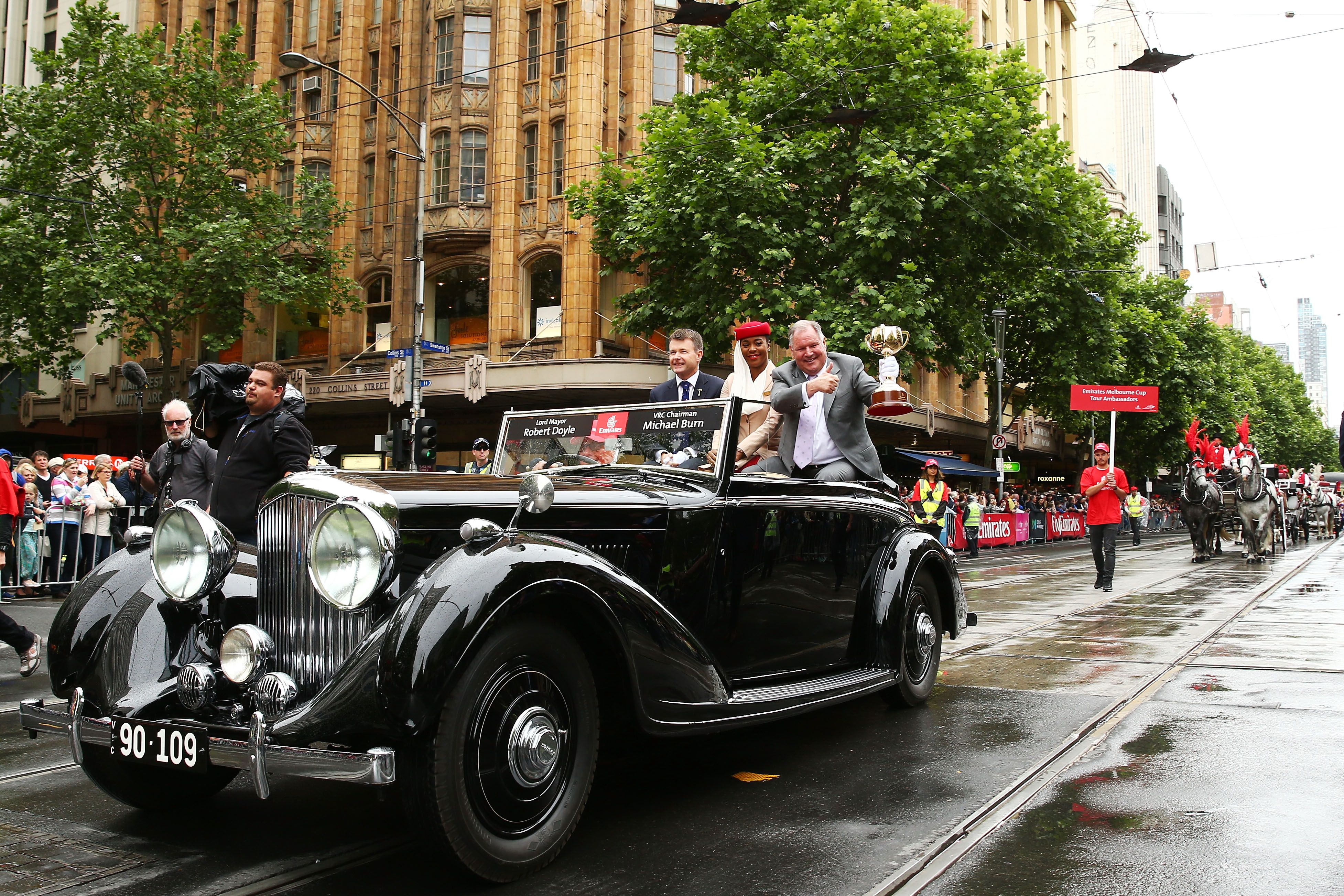}
\caption{A query image from the public test set, showing jockey Michelle Payne's victory parade.}
\Description{Following her historic Melbourne Cup win aboard Prince of Penzance, jockey Michelle Payne and connections are driven in a vintage car along Collins Street in Melbourne on November 3, 2015.}
\label{fig:query_img}
\end{figure}

We present a qualitative analysis of a sample from the public test set. Figure \ref{fig:query_img} shows an image from the dataset. Below, we compare a caption generated by a baseline model (which only has access to the image) with the caption generated by our full ZSE-Cap pipeline.

Baseline Caption (Image-Only):

\begin{quote}
\small 

The image captures a jubilant, rain-slicked street scene in Melbourne, Australia, likely during a daytime parade, judging by the overcast lighting and the presence of a large crowd. The dominant color palette is a muted, cool gray-blue, punctuated by the dark sheen of the vintage car and the vibrant red of the crowd’s attire. A soft, diffused light filters through the tall buildings, creating a slightly melancholic yet celebratory atmosphere.

\end{quote}

ZSE-Cap Caption (Image + Article + Prompt):

\begin{quote}
\small 

Following her historic Melbourne Cup win aboard Prince of Penzance, jockey Michelle Payne and connections are driven in a vintage car along Collins Street in Melbourne on November 3, 2015. Payne’s victory marked the first time a female jockey had won the prestigious race, a moment celebrated as a significant step for women in the traditionally male-dominated sport of horse racing – a point Payne emphasized in her post-race comments, challenging those who doubted women’s capabilities. Visible in the car with Payne are, from left, Robert Doyle (Lord Mayor of Melbourne) and Michael Burn (Victoria Racing Club Chairman), acknowledging the significance of the event which drew congratulations from figures including Australian Prime Minister Malcolm Turnbull. The win, highlighted by Payne wearing the colors of the Suffragettes, resonated as a triumph of determination and equality.

\end{quote}

The comparison reveals a stark difference in contextual understanding. The baseline caption accurately describes the visible scene but completely fails to capture the event's essence. The description is generic and devoid of any real-world meaning.

In contrast, our ZSE-Cap caption successfully grounds the image in its specific context. By retrieving the correct news article and using our structured prompt, the system identifies the key figure, the specific event, the location, and the profound narrative behind the image. This demonstrates a clear shift from simple visual description to true, event-level comprehension, which is the core challenge of this task.

\subsection{Error analysis}

We identified two primary scenarios where retrieval accuracy was compromised.

\begin{itemize}
    \item Contextual ambiguity from near-duplicate images: The most significant challenge arises when a single image, or a near-duplicate, is associated with multiple news articles in the database. This often occurs with stock photos or widely syndicated images of prominent figures (e.g., a politician at a podium). Our retrieval module, which is fundamentally based on matching visual features via L2 distance, struggles to disambiguate in these cases. Since the visual embeddings for the identical images are nearly the same, the ensemble score provides no strong signal to select the one linked to the correct event narrative. The choice of which article's context to use for captioning can become arbitrary, leading to a correct image retrieval but a contextually incorrect caption. This highlights a core limitation of a purely image-to-image retrieval pipeline for tasks where context is paramount and not fully encapsulated within the image pixels.

    \item Sensitivity to severe visual perturbations: A second limitation is the system's reduced accuracy when faced with query images that have undergone heavy augmentation or contain significant noise. Although the foundation models (CLIP, SigLIP, DINOv2) are generally robust, we observed a noticeable drop in retrieval performance with aggressive transformations (e.g., extreme cropping, severe compression artifacts, or drastic color shifts). These perturbations can alter the feature vectors extracted by the encoders sufficiently to increase the L2 distance to their pristine counterparts in the database. As a result, the correct image may be ranked lower than visually dissimilar but cleaner candidates, leading to a retrieval failure. This indicates that while effective, the robustness of current vision models still has boundaries, a practical consideration for real-world applications where image quality can be highly variable.
\end{itemize}

Analyzing these failure points underscores that the primary hurdles are not in general visual understanding but in disambiguating context and maintaining robustness under extreme conditions. Future work should directly address these challenges, perhaps by incorporating textual signals earlier in the retrieval process or by exploring methods to enhance feature invariance to severe perturbations.

\section{Conclusion}

In this paper, we presented ZSE-Cap, an effective zero-shot system for the EVENTA shared task on article-grounded image retrieval and captioning. Our approach tackles the two coupled sub-tasks without requiring any training or fine-tuning on the competition's data.

Our primary contributions are:

\begin{itemize}
    \item A robust image retrieval module based on an ensemble of diverse vision models (CLIP, SigLIP, and DINOv2), which demonstrated high accuracy in locating relevant images within a large-scale database.

    \item An advanced captioning pipeline that uses prompt engineering to steer a Large Language Model (Gemma 3) to generate contextually rich captions that deeply connect an article's narrative with visual details.
\end{itemize}

Our system not only achieved a top-4 rank, validating the effectiveness of our approach, but more importantly, it highlights a promising direction for future research. It demonstrates that complex and effective multimodal systems can be built by intelligently composing powerful, pre-existing foundation models, rather than creating specialized models from scratch.

Future work could explore automated prompt optimization or more dynamic model fusion techniques. We believe ZSE-Cap serves as a clear testament to the significant potential of zero-shot methods in solving increasingly complex, real-world multimodal challenges.

\begin{acks}
We would like to thank SELAB @ HCMUS for providing the OpenEvents V1 dataset and organizing the EVENTA grand challenge.
\end{acks}
% \input{content/instruction}

\bibliographystyle{ACM-Reference-Format}
\bibliography{references}

\end{document}